\documentclass[11pt]{article}

\usepackage[final]{acl}

\usepackage{times}
\usepackage{latexsym}

\usepackage[T1]{fontenc}

\usepackage[utf8]{inputenc}

\usepackage{microtype}

\usepackage{inconsolata}

\usepackage{graphicx}

\usepackage{tabularx,colortbl,booktabs,makecell,multirow,amssymb,url,cleveref,pifont,tcolorbox,CJKutf8}

\title{Benchmarking Large Vision-Language Models on CFMME: A Comprehensive Chinese Financial Multimodal Evaluation Dataset}

\author{
 \textbf{Qian Chen\textsuperscript{1}},
 \textbf{Xianyin Zhang\textsuperscript{1}},
 \textbf{Yanzhi Liu\textsuperscript{1}},
\\
 \textbf{Lifan Guo\textsuperscript{1}},
 \textbf{Feng Chen\textsuperscript{1}},
 \textbf{Chi Zhang\textsuperscript{1}}
\\
\\
 \textsuperscript{1}Qwen DianJin Team, Alibaba Cloud Computing
\\
 \small{
   \{\href{xiaoyu.cq@alibaba-inc.com}{xiaoyu.cq}, \href{zhangxianyin.zxy@alibaba-inc.com}{zhangxianyin.zxy}, \href{xiandou.lyz@alibaba-inc.com}{xiandou.lyz}, \href{lifan.lg@alibaba-inc.com}{lifan.lg}, \href{betterman.chenf@alibaba-inc.com}{betterman.chenf}, \href{edward.zhang@alibaba-inc.com}{edward.zhang}\}@alibaba-inc.com
 }
}

\begin{document}
\maketitle
\begin{abstract}
The emergence of Large Vision-Language Models (LVLMs) has substantially expanded model capabilities beyond text-only understanding, enabling unified inference across both visual and textual modalities and supporting a broader range of real-world applications. To comprehensively evaluate the perception, understanding, reasoning, and cognition capabilities of LVLMs throughout the entire financial business workflow in Chinese contexts, we introduce CFMME, a novel Chinese financial multimodal evaluation benchmark. CFMME comprises 6,052 instances spanning from fundamental academic knowledge to complex real-world applications, covering eight primary financial image modalities and four core multimodal tasks. On CFMME, we conduct a thorough evaluation of representative LVLMs. The results show that the state-of-the-art model attains an overall accuracy of 66.11\% on the question answering task and an average score of 77.18 on the detection, recognition, and information extraction tasks, indicating substantial room for improvement in current LVLMs. In addition, we conduct detailed analyses of error causes, cross-modal capabilities, and multi-orientation settings, yielding valuable insights for future research. We hope that CFMME will spur further progress in LVLMs, especially by improving their performance on multiple multimodal tasks in the financial domain.
\end{abstract}

\section{Introduction}
The remarkable capabilities demonstrated by Large Vision-Language Models (LVLMs) have brought significant advances beyond pure text understanding, extending to tasks that require joint understanding and reasoning over images and text. In order to comprehensively evaluate the capabilities of LVLMs, new benchmarks have been introduced to assess a varied range of abilities exhibited by these models. Notable examples including MME \cite{fu2023mme}, MMMU \citep{yue2024mmmu}, and MMBench \citep{liu2024mmbench} are widely employed to evaluate LVLMs’ capabilities in English. Meanwhile, benchmarks such as CMMU \citep{he2024cmmu} and CMMMU \citep{zhang2024cmmmu} are widely used to assess their capabilities in Chinese.

\begin{table}[t]
    \centering
    \footnotesize
    \resizebox{\linewidth}{!}{
        \begin{tabular}{lcccccc}
            \toprule
            \multirow{2}{*}{\textbf{Benchmarks}} & \multirow{2}{*}{\textbf{Language}} & \multirow{2}{*}{\textbf{\makecell{Primary \\ Source}}} & \multicolumn{4}{c}{\textbf{Task}} \\
            \cmidrule{4-7}
            & & & \textbf{\makecell{Question \\ Answering}} & \textbf{Detection} & \textbf{Recognition} & \textbf{\makecell{Information \\ Extraction}} \\
            \midrule
            FAMMA & EN & \makecell{Textbooks \\ Examinations} & \checkmark & & & \\
            \midrule
            FinMME & EN & \makecell{Research Reports \\ Webpage Screenshots} & \checkmark & & & \\
            \midrule
            MME-Finance & EN/CN & \makecell{Research Reports \\ Applications} & \checkmark & & & \\
            \midrule
            VisFinEval & CN & \makecell{Research Reports \\ Annual Reports \\ Examinations} & \checkmark & & & \\
            \midrule
            CFMME & CN & ALL & \checkmark & \checkmark & \checkmark & \checkmark \\
            \bottomrule
        \end{tabular}
    }
    \caption{Comparison of various existing financial multimodal benchmarks.}
    \label{tab:comparison}
\end{table}

\begin{figure*}
    \centering
    \includegraphics[width=0.9\textwidth]{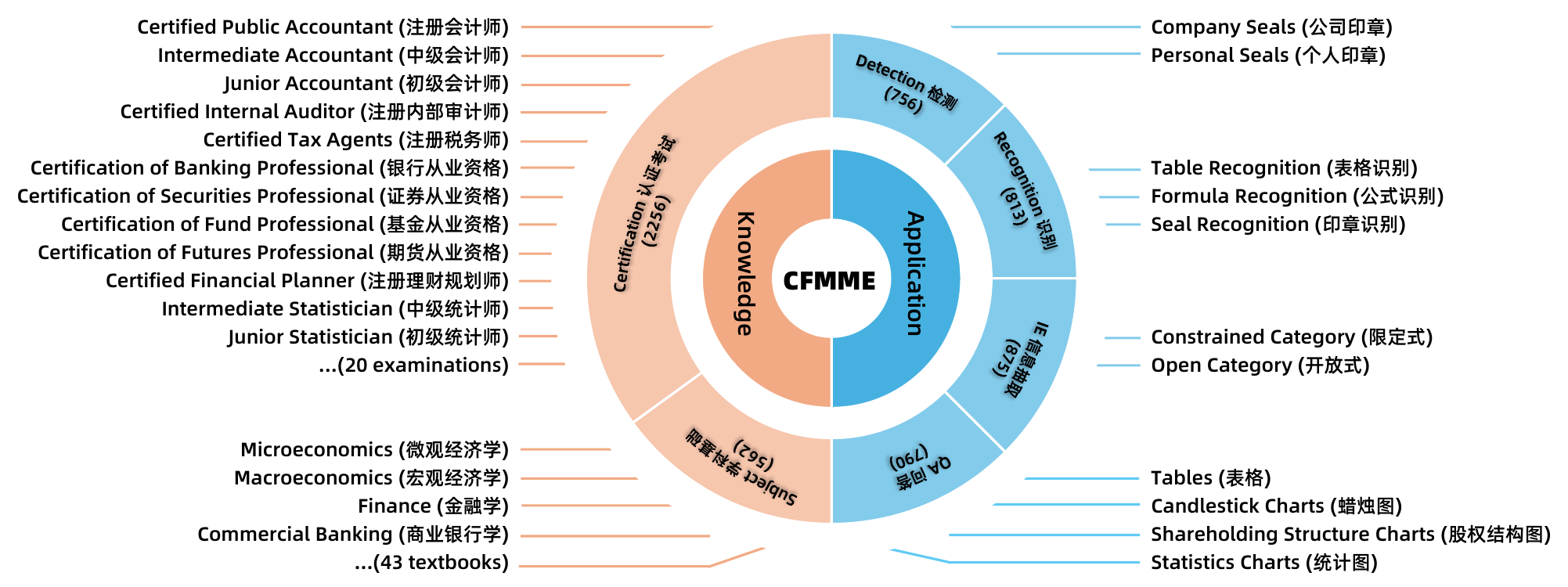}
    \caption{Overview diagram of our benchmark.}
    \label{fig:overview}
\end{figure*}

\begin{table*}[h]
    \centering
    \footnotesize
    \resizebox{\linewidth}{!}{
    \begin{tabularx}{\textwidth}{l|l|c|X}
        \toprule
        \textbf{Category} & \textbf{Subcategory / Task} & \textbf{Size} & \textbf{Description} \\
        \midrule
        \multirow{2}{*}{Knowledge} & Subject & 562 & 43 textbooks across diverse subjects \\
        & Certification & 2,256 & 20 types of qualification certification examinations \\
        \midrule
        \multirow{6}{*}{Application} & Seal Detection & 756 & company seals and personal seals \\
        & Information Extraction & 875 & constrained-category and open-category \\
        & Seal Recognition & 213 & shapes of oval, round, square, triangle and rhombus \\
        & Table Recognition & 400 & table in a single page and table across two pages\\
        & Formula Recognition & 200 & formula with pure symbols and formula with Chinese characters \\
        & Question Answering & 790 & tables, candlestick charts, shareholding structure charts and statistic charts \\
        \midrule
        Total & - & 6,052 & - \\
        \bottomrule
    \end{tabularx}
    }
    \caption{Statistics of our benchmark.}
    \label{tab:statistics}
\end{table*}

In the financial domain, practitioners often contend with a wide variety of visual data, including tables, charts, seals, and diverse document types. Effectively processing and interpreting such data is essential to the entire financial workflow. However, assessing multimodal capabilities in financial scenarios, whether in Chinese or English contexts, has not been fully explored, with notable limitations remaining. Specifically, the English benchmarks FAMMA \citep{xue2024famma} and FinMME \citep{luo2025finmme} offer valuable financial multimodal evaluation, yet their coverage is relatively narrow: FAMMA is primarily drawn from textbooks and examinations, while FinMME largely relies on research reports and webpage screenshots. Meanwhile, the bilingual benchmark MME-Finance \citep{gan2025mme} is constrained by limited data volume. In Chinese, VisFinEval \citep{liu2025visfineval} concentrates exclusively on question answering tasks, providing limited support for a more comprehensive assessment spanning broader multimodal tasks. As shown in Table \ref{tab:comparison}, resent benchmarks have limited data source and multimodal task coverage, which may result in fragmented and sometimes misleading insights. No existing benchmarks systematically and comprehensively assess perception, understanding, reasoning, and cognition capabilities of LVLMs across the full spectrum of financial business workflow using extensive data sources and task designs.

Inspired by CFLUE \citep{zhu2024benchmarking} benchmark, which provides a comprehensive benchmark for evaluating through various natural language processing (NLP) tasks in Chinese, we introduce a novel benchmark named CFMME (Chinese Financial Multimodal Evaluation), which addresses the aforementioned challenges by comprehensively evaluating LVLMs through various data sources, image types, and multimodal tasks. Specifically, CFMME has the following key features: \textbf{\ding{172} Extensive Data Coverage:} It is designed to assess both academic knowledge and practical application. For the knowledge assessment, one subset is complied from 43 textbooks across diverse subjects such as Microeconomics, Macroeconomics, and Finance. The other subset is derived from 20 types of certification examinations, which covers a wide range of difficulty levels such as Junior Accountant and Intermediate Accountant, and spans various certified domains such as Certified Tax Agent and Certified Financial Planner. For the application assessment, based on real-world financial scenarios, data are sourced from financial websites, applications, annual reports, research reports, invoices and other visually-rich documents. \textbf{\ding{173} Rich Image Modalities:} It includes eight primary image types--tables, seals, formulas, candlestick charts, shareholding structure charts, statistical charts, documents, and others. Furthermore, these images are presented in diverse formats to reflect real-world complexities, such as blurred backgrounds, skewed layouts, carrying watermarks, captured by mobile phones, and containing handwritten text. \textbf{\ding{174} Diverse Multimodal Tasks:} It covers four core multimodal tasks: question answering, detection/grounding, recognition, and information extraction, comprehensively assessing the perception, understanding, reasoning, and cognition capabilities of LVLMs. The recognition task is further decomposed into three sub-tasks: seal, table, and formula recognition. 

The overview and statistics of CFMME are presented in Figure \ref{fig:overview} and Table \ref{tab:statistics}. In summary, the contributions of our work are three-fold:
\begin{itemize}
    \item We propose a comprehensive Chinese financial multimodal evaluation benchmark called CFMME with 6,052 instances, spanning from fundamental academic knowledge to complex real-world applications, covering eight primary financial image types and four core multimodal tasks.
    \item We conduct a thorough evaluation of 14 different LVLMs on CFMME in a zero-shot setting, providing a quantitative analysis of their strengths and weaknesses.
    \item In-depth analysis reveals valuable insights and guidance for enhancing LVLMs to achieve advanced levels of multimodal capabilities in the financial domain.
\end{itemize}

\section{Related Works}

\subsection{Text-Only Financial Benchmarks}
Recent advancements in Large Language Models (LLMs) have demonstrated exceptional capabilities across various complex tasks, including MMLU \citep{hendryckstest2021}, GPQA \citep{rein2024gpqa} and others. In the financial domain, FLUE \citep{shah2022flue} utilizes datasets from existing literature to establish a suite of benchmarks across five NLP tasks. FinanceBench \cite{islam2023financebench}, evaluates the performance on open book financial question answering. FinEval \citep{guo2025fineval} is designed to evaluate financial domain knowledge and practical abilities, categorized into four different key areas. CFLUE \citep{zhu2024benchmarking} provides Chinese datasets across distinct groups of NLP tasks, and is tailored for both knowledge assessment and application assessment. However, these text-only financial benchmarks assess essential language understanding capabilities while overlooking the visual information that often underpins real-world decision making.
\subsection{Multimodal Financial Benchmarks}
Recent advancements in LLMs have driven notable progress in LVLMs, enabling exceptional capabilities for unified visual-linguistic understanding. General-purpose multimodal benchmarks, such as MME \citep{fu2023mme}, MMMU \citep{yue2024mmmu}, and MMBench \citep{liu2024mmbench}, have been introduced to assess the broad capabilities of LVLMs. In the financial domain, FAMMA \citep{xue2024famma} provides a benchmark for financial question-answering tasks in English, though its data primarily derives from textbooks and examinations. Similarly, FinMME \citep{luo2025finmme} offers a robust evaluation framework;however, its data mainly originates from research reports and webpage screenshots. MME-Finance \citep{gan2025mme} introduces a bilingual visual question-answering benchmark but is constrained by its limited data volume. Meanwhile, VisFinEval \citep{liu2025visfineval} consists of large-scale annotated question–answer pairs structured across three hierarchical scenario depths, yet it lacks support for multi-task evaluations that better represent the comprehensive multimodal characteristics of LVLMs.

\section{CFMME: Chinese Financial Multimodal Evaluation}
We propose CFMME, a multimodal benchmark tailored to Chinese financial domain, evaluating the proficiency of LVLMs from both knowledge and application perspective. This section provides a comprehensive introduction to CFMME, including detailed data construction (Section \ref{sec:ka} and \ref{sec:aa}) and quality control protocols (Section \ref{sec:qc}). 

\subsection{Knowledge Assessment}
\label{sec:ka}
According to the different data sources, we divide this assessment into two subsets: subject and certification. The subject subset is sourced from 43 publicly available textbooks, encompassing a wide range of disciplines. Similarly, the certification subset is derived from publicly accessible and challenging qualification certification examinations designed to evaluate real-world expertise and problem-solving skills. These examinations span 20 distinct types, representing various difficulty levels and certified domains. Additionally, we observe that CFLUE \citep{zhu2024benchmarking} contains several similar examinations and certain instances with <img src></img> tags. Therefore, we extract the images by downloading them from the src attribute of these tags and incorporating them into our benchmark.

The original data are sourced from PDFs, Microsoft Word documents and webpage available on the Internet. We first use tools such as pdfplumber\footnote{\url{https://github.com/jsvine/pdfplumber}} and PaddleOCR\footnote{\url{https://github.com/PaddlePaddle/PaddleOCR}} to parse these documents, and then extract questions, options, answers and explanations one by one. Finally, we remove HTML tags from the text by regular expressions, correct incorrect characters through manual annotation, and filter out all text-only instances.

After processing and annotation, we compile a total of 2,818 instances paired with corresponding images, including 562 instances from the subject subset and 2,256 instances from the certification subset. Among them, there are multiple-choice questions and true-or-false questions, with the true-or-false questions presented in multiple-choice format. To assess the model's capability to understand and reason both single and multiple images, this assessment contains 308 instances with multiple images and 2,510 instances with one single image. In addition, in order to increase the difficulty of answering questions, this assessment has a maximum of five options, and consists of 560 multi-answers questions and 2,258 single-answer questions. Unlike traditional text-only benchmarks and other multimodal benchmarks where both questions and options are solely provided in textual format, we deliberately introduce greater complexity and diversity by taking screenshots of the options for some questions as one or multiple images during the data processing stage. This intentional approach enriches the benchmark's multimodal nature and adds an additional layer of challenge. Specifically, this assessment comprises 238 questions with options presented in visual format, alongside 2,580 questions with options provided in textual format.

\subsection{Application Assessment}
\label{sec:aa}
To thoroughly evaluate the capabilities of LVLMs in the financial domain, it is inadequate to rely solely on questions derived from textbooks and examinations, as establishing a more comprehensive, multidimensional evaluation framework should also capture their real-world performance and practical utility. Accordingly, we collect extensive real-world data from diverse sources, including annual reports, research reports, and other financial documents, as well as financial websites and applications. The ability to effectively perceive and interpret these images is critical for downstream decision-making and constitutes a key component of financial workflows such as underwriting, investment research, and investment advisory. Considering common multimodal tasks in financial workflows, we mainly assess the performance of LVLMs across four tasks in this assessment: detection, recognition, information extraction, and question answering.

\subsubsection{Detection}
Locating objects in an image is commonly referred to as object detection or visual grounding. In financial documents, accurately locating seals—which are ubiquitous—is a prerequisite for subsequent seal recognition or seal removal. Therefore, we construct seal detection task by collecting images from documents with seals and prompting LVLMs to output both the bounding box and the category of the seal (divided into company seals and personal seals). 

During annotating, we first use our internally trained seal detection model to pre-label the collected data, and then use label studio\footnote{\url{https://labelstud.io/}} for verification. Finally, we collect a total of 756 images, including 528 images that contain both two types of seals at the same time and 228 images that only contain one type.

\subsubsection{Recognition}
Accurate recognition of financial documents is essential for extracting critical information and enabling reliable interpretation. This assessment focuses on three representative and challenging element types commonly found in financial documents: seals, tables, and formulas. Seals are frequently degraded and overlap with other content, and they often contain curved or warped text and suffer from strong background interference. Recognizing tables requires recovering the complete row–column structure, including merged cells, under noisy, broken, or even missing ruling lines. Recognizing formulas that are inherently two-dimensional demands parsing spatial relationships (fractions, superscripts, subscripts, and matrices) into correct LaTeX, rather than merely recognizing individual symbols.

\paragraph{Seal Recognition.} We select data from ICDAR 2023 competition on reading the seal title (ReST) \citep{yu2023rest}. To minimize the risk of data contamination, we only select instances from test sets, for which the competition organizers did not release annotations. Additionally, this competition only requires recognizing the seal title rather than the entire seal content, so we re-annotate this seals. After annotation, we collect a total of 206 seal images, whose shapes span a range of types, including oval, circular, square, triangular, and rhombic.
\paragraph{Table Recognition.} We select data from ICDAR 2023 Competition on Structured Text Extraction from Visually-Rich Document Images (SVRD) \citep{yu2023svrd}, TabRecSet\citep{yang2023large}, and in-house dataset. SVRD dataset contains more than 50 types of visually-rich document images, from which we extract tables from receipts, bills, forms, and purchase orders. TabRecSet dataset is a large-scale bi-lingual dataset for end-to-end table recognition, from which we extract tables from invoices. For in-house dataset, we extract tables from annual reports and research reports. After processing and annotation, we collect a total of 400 instances, among which 312 instances contain tables in one single page and 82 instances contains tables across two consecutive pages. To ensure diversity, our table images includes various types, such as full-line, incomplete-line, and no-line.
\paragraph{Formula Recognition.} We select data from aforementioned textbooks and examinations. After processing and annotation, we compile 200 instances, of which 128 are formulas composed solely of mathematical symbols, whereas 72 contain symbols and Chinese characters.

We design an innovative annotation methodology. Specifically, seals, tables and formulas are first located using our internal layout analysis model. Then, for the cropped images, two state-of-the-art models, i.e., Qwen3-VL-Plus \citep{Qwen3-VL} and MinerU2.5 \citep{niu2025mineru2} are used to generate pseudo-labels. When the edit distance between two models' results is large, we use DianJin-OCR-R1 \cite{chen2025dianjin} to refine. We tend to choose the instance with a larger edit distance because the inconsistency to some extent represents the difficulty of this instance. Finally, all annotations undergo rigorous verification by human annotators.

\subsubsection{Information Extraction}
Information extraction, which aims to jointly perform optical character recognition (OCR) and key–value pair extraction within a unified framework, has garnered increasing attention due to its pivotal role in understanding financial visually-rich documents—such as receipts, invoices, bills, bank cards, and account statements—thereby enabling automated reconciliation, compliance verification, and risk control. Currently, this task can be divided into two categories: constrained-category and open-category datasets, depending on whether the key is predefined. For the constrained-category, the target key to be extracted is given.

We select data from SVRD and CMB2017\footnote{\url{https://www.heywhale.com/mw/dataset/5954cf1372ead054a5e25870}}. SVRD dataset has two subsets: HUST-CELL and Baidu-FEST, while CMB2017 dataset contains fake cards from China Merchants Bank and other banks. For the constrained-category, data are derived from Baidu-FEST and CMB2017, whereas data for open-category are derived from HUST-CELL.

The original annotation of these datasets designed to extract entities and entity relationships is no longer applicable, so we re-annotate these datasets using a standardized JSON format. To minimize the risk of data contamination, we only select instances from test sets. Meanwile, to ensure diversity, we exclusively select images that are presented in diverse formats such as blurred backgrounds, skewed layouts, carrying watermarks, captured by mobile phones, and containing handwritten text. After processing and annotation, we finally collect 500 instances from HUST-CELL, 275 instances from Baidu-FEST, and 100 instances from CMB2017.

\subsubsection{Question Answering}
In financial applications, question answering extends beyond generic scene understanding to interpreting domain-specific artifacts such as candlestick charts and shareholding structure charts. LVLMs must accurately extract and reason over visual evidence—such as numerical values, axis scales, and legends—while grounding responses in financial concepts. For this assessment, we construct a dataset of financial images by combining manual curation with automated web crawling from financial documents, webpages, and application screenshots. To ensure broad coverage, the dataset includes multiple image types: tables, candlestick charts, shareholding structure charts and statistics charts (including line charts, bar charts, pie charts, and mixed charts).

We employ a multi-stage QA generation pipeline: we first utilize Qwen3-VL-Plus \citep{Qwen3-VL} to generate 10 candidate questions for every image. Then, human experts choose questions from these candidate questions and decide whether to increase their difficulty. In the annotation stage, we employ a parallel annotation method where both human annotators and LVLMs independently to generate preliminary answers based on questions and images. Finally, we adopt a consistency-based review workflow: if human annotators and LVLMs agree, a single expert conducts a quick validation; if they diverge, three experts perform an in-depth assessment to establish the final ground truth.

After annotation, we compile 790 instances including 505 instances in multiple-choice format and 285 instances in open-ended format (with brief answers), which evaluates LVLMs' ability to perform cross-modal understanding, temporal sequence recognition, horizontal comparisons, numerical calculation, causal analysis, etc.

\begin{table*}[h]
    \centering
    \footnotesize
        \begin{tabular}{l|c|cc|c}
            \toprule
            \multirow{2}{*}{\textbf{Models}} & \multirow{2}{*}{\textbf{Avg.}} & \multicolumn{2}{c|}{\textbf{Knowledge}} & \textbf{Application} \\
            & & Acc. & F1 & Acc. \\
            \midrule
            Kimi-VL-A3B-Instruct & 27.97 & 36.52±0.57 & 39.37±0.63 & 19.41±0.64 \\
            Kimi-VL-A3B-Thinking & 41.85 & 47.33±0.52 & 48.24±0.60 & 36.37±1.05 \\
            ERNIE-4.5-VL-28B-A3B & 45.68 & 49.28±0.31 & 49.61±0.31 & 42.07±0.64 \\
            ERNIE-4.5-VL-28B-A3B-Thinking & 52.70 & 59.40±0.26 & 63.38±0.22 & 45.99±0.74  \\
            GLM-4.1V-9B-Thinking & 51.51 & 61.21±0.09 & 62.27±0.20 & 41.81±0.82 \\
            GLM-4.5V (non-thinking) & 57.52 & 64.90±0.09 & 65.42±0.06 & 50.13±0.46 \\
            GLM-4.5V (thinking) & 60.78 & 70.20±0.32 & 72.72±0.17 & 51.35±0.70 \\
            InternVL3.5-8B (non-thinking) & 46.07 & 53.35±0.24 & 52.82±0.26 & 38.78±0.19 \\
            InternVL3.5-8B (thinking) & 50.88 & 58.47±0.49 & 62.37±0.47 & 43.29±0.00 \\
            InternVL3.5-241B-A28B (non-thinking) & 53.29 & 64.05±0.21 & 62.97±0.22 & 42.53±0.91 \\
            InternVL3.5-241B-A28B (thinking) & 59.90 & 69.72±0.30 & 71.01±0.37 & 50.08±0.60 \\
            Qwen3-VL-8B-Instruct & 52.83 & 61.85±0.48 & 63.25±0.47 & 43.80±0.70 \\
            Qwen3-VL-8B-Thinking & 53.85 & 63.47±0.26 & 66.28±0.48 & 44.22±0.41 \\
            Qwen3-VL-235B-A22B-Instruct & \underline{63.31} & \underline{74.05±0.45} & \underline{73.82±0.42} & \underline{52.57±0.73} \\
            Qwen3-VL-235B-A22B-Thinking & \textbf{66.11} & \textbf{77.27±0.59} & \textbf{77.88±0.14} & \textbf{54.94±0.51} \\
            \midrule
            Llama-3.2-11B-Vision-Instruct & 20.20 & 21.79±0.38 & 23.49±0.39 & 18.61±0.77 \\
            GPT-4o & 46.96 & 56.87±0.90 & 58.51±0.97 & 37.05±0.70 \\
            \midrule
        \end{tabular}
    \caption{Overall comparison on question answering tasks. Scores in \textbf{bold} and \underline{underline} denote the top and second-best performances. When calculating the overall average score, we use the accuracy of each assessment.}
    \label{tab:qa}
\end{table*}

\begin{table*}
    \centering
    \resizebox{\textwidth}{!}{
        \begin{tabular}{l|c|c|cc|cc|c|c}
            \toprule
            \multirow{3}{*}{\textbf{Models}} & \multirow{3}{*}{\textbf{Avg.}} & \multicolumn{5}{c|}{\textbf{Rec}} & \textbf{Det} & \textbf{IE} \\
            & & \multicolumn{1}{c}{\textbf{Seal}} & \multicolumn{2}{c}{\textbf{Table}} & \multicolumn{2}{c|}{\textbf{Formula}} & & \\
            & & 1-NED & TEDS & STEDS & CDM & ExpRate@CDM & mAP & F1 \\
            \midrule
            Kimi-VL-A3B-Instruct & 50.39 & 50.86±1.34 & 54.20±1.04 & 63.31±0.81 & 71.44±0.14 & 15.83±1.18 & 0.00±0.00 & 75.46±1.29 \\
            ERNIE-4.5-VL-28B-A3B & 56.61 & 73.86±1.20 & 57.49±0.50 & 66.91±0.70 & 70.98±0.98 & 11.67±0.62 & 0.00±0.00 & 80.71±0.39 \\
            GLM-4.5V (non-thinking) & \underline{71.77} & 77.25±0.16 & \underline{71.89±0.05} & 75.91±0.07 & 83.24±0.16 & 17.00±0.00 & \underline{37.45±0.11} & \underline{89.03±0.75} \\
            InternVL3.5-8B (non-thinking) & 59.03 & 65.70±0.46 & 59.96±0.48 & 68.67±0.52 & \underline{89.02±0.79} & 31.17±0.85 & 0.19±0.04 & 80.29±0.62 \\
            InternVL3.5-241B-A28B (non-thinking) & 59.49 & 60.25±1.01 & 65.60±0.27 & 72.71±0.23 & 88.06±0.52 & 27.33±0.47 & 0.42±0.06 & 83.10±0.20 \\
            Qwen3-VL-8B-Instruct & 71.04 & \underline{79.15±0.08} & 71.35±0.13 & \underline{76.60±0.09} & 87.57±0.37 & \underline{36.50±0.00} & 28.27±0.15 & 88.86±0.38 \\
            Qwen3-VL-235B-A22B-Instruct & \textbf{77.18} & \textbf{88.27±0.28} & \textbf{78.92±0.24} & \textbf{81.61±0.26} & \textbf{95.01±0.14} & \textbf{49.17±0.47} & \underline{30.27±0.12} & \textbf{93.43±0.34} \\
            \midrule
            Llama-3.2-11B-Vision-Instruct & 27.42 & 13.44±1.13 & 28.92±0.21 & 42.46±0.19 & 47.23±0.44 & 2.67±0.24 & 0.00±0.00 & 47.52±0.38 \\
            GPT-4o & 57.89 & 81.45±1.17 & 57.11±0.50 & 66.02±0.73 & 83.73±1.16 & 20.5±0.41 & 0.30±0.21 & 66.88±0.28 \\
            \bottomrule
        \end{tabular}
    }
    \caption{Overall comparison on recognition, detection and information extraction tasks. Scores in \textbf{bold} and \underline{underline} denote the top and second-best performances. Non-reasoning models are tested on these tasks. When calculating the overall average score, we use 1-NED, TEDS, CDM, mAP and F1 for corresponding tasks.}
    \label{tab:ocr}
\end{table*}

\subsection{Quality Control}
\label{sec:qc}
To ensure high-quality annotations, we recruit a team of 20 annotators. For the knowledge assessment, three PhDs in Economics and Business Administration are employed to annotate the subject subset. For the certification subset, we engage eight experts who hold multiple professional qualification certifications. For each examination, all questions, answers, and explanations are independently reviewed and verified by three experts who have passed that examination. For the application assessment, we recruit 12 annotators, consisting of six junior annotators, three PhDs (as mentioned above) and three analysts. Six junior annotators annotate the detection, recognition and information extraction data, while three PhDs and three analysts annotate the question answering data. All of them undergo training for a week and access guidance documents. The whole annotation and review process took approximately 1600 hours.

\section{Experiment}
\subsection{Settings}
\paragraph{Models.} We conduct experiments across a diverse range of model architectures and parameter scales, with close-source models accessed through their respective APIs and open-source models deployed locally by vLLM\footnote{\url{https://github.com/vllm-project/vllm}} on 8 NVIDIA A100 GPUs. The evaluated models include GPT-4o\footnote{\url{https://openai.com/index/hello-gpt-4o/}}, Llama-3.2-Vision\footnote{\url{https://www.llama.com/models/llama-3/}}, Qwen3-VL Series \citep{Qwen3-VL}, InternVL3.5 Series \citep{wang2025internvl3}, GLM-V Series \citep{vteam2025glm45}, Kimi-VL Series \citep{team2025kimi}, and ERNIE-4.5-VL Series \citep{ernie2025technicalreport}. More details are in Appendix \ref{app:model}.
\paragraph{Metrics.} In the knowledge assessment, we measure the performance using accuracy and weighed F1 score. In the application assessment, we report Mean Average Precision (mAP) for detection, 1-Normalized Edit Distance (1-NED) \citep{zhang2019icdar} for seal recognition, Character Detection Matching (CDM) and ExpRate@CDM \citep{wang2024cdm} for formula recognition, Tree-Edit-Distance-based Similarity (TEDS) and Structure TEDS \citep{zhong2020image} for table recognition, field-level F1 score \citep{hwang2019post, kim2022ocr} for information extraction, and accuracy for question answering.
\paragraph{Prompts.} Our evaluation of LVLMs is conducted in a zero-shot setting. We use a single prompt for each task, maintaining consistency by employing the same prompt within each group. Details can be found in Appendix \ref{app:prompt}.
\paragraph{Parameters.} For parameters such as temperature and top\_p, we adopt the recommended settings mentioned in each model’s paper or website to elicit the best performance from each model. Details can be found in Appendix \ref{app:para}.

\subsection{Main Results}
As shown in Table \ref{tab:qa}, Qwen3-VL-235B-A22B-Thinking achieves the highest average score 66.11\% in the question answering task, followed by Qwen3-VL-235B-A22B-Instruct. Even the top model fails to reach 70\% on our benchmark, whereas in real-world settings an accuracy of over 80\% is typically required for deployment in production. In particular, on the question answering task in the application assessment, demands exceptionally strong reasoning and cognition abilities, as well as a deep understanding of financial terminology and its flexible use, the highest score is only 54.94\%. These findings suggest, on the one hand, that our benchmark is highly challenging and can comprehensively evaluate the reasoning and cognition capabilities of LVLMs; on the other hand, they also indicate that the performance of LVLMs in the financial domain remains to be improved. In addition, as CFMME is tailored to the Chinese context, English-oriented models such as GPT-4o perform worse than Chinese-oriented models, which is also consistent with common expectations.

As shown in Table \ref{tab:ocr}, Qwen3-VL-235B-A22B-Thinking emerges as the top performer with an average score of 77.18 on the detection, recognition and information extraction tasks, which mainly assess the perception and understanding capabilities of LVLMs. It is worth noting that existing models perform well on the recognition and information extraction task, yet they still exhibit significant shortcomings on the object detection task. Aside from Qwen3-VL series and GLM-4.5V, most models struggle to accurately localize objects. Even for the best-performing GLM-4.5V, its mAP is only 37.45, which demonstrate that fine-grained grounding remains a major challenge for current LVLMs.

We conduct three rounds of testing, and the standard deviations across all evaluation dimensions remain consistently below 1.5\%, which indicates the high reliability and reproducibility of our evaluation framework and also confirming the consistency of model behaviors across multiple test rounds.

\subsection{Analysis}
\begin{figure}
\begin{tcolorbox}[title=Error Cases, colback=gray!5, colframe=black, fonttitle=\bfseries]

    \begin{minipage}[h]{\linewidth}
    \centering
    \includegraphics[width=0.8\linewidth]{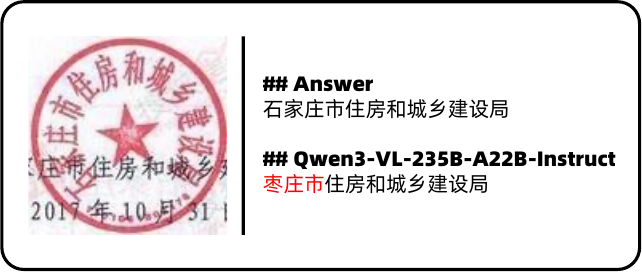}
    \end{minipage} 

    \begin{minipage}[h]{\linewidth}
    \centering
    \includegraphics[width=0.8\linewidth]{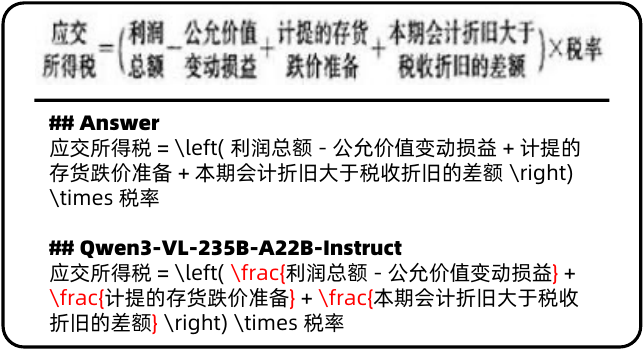}
    \end{minipage} 

    \begin{minipage}[h]{\linewidth}
    \centering
    \includegraphics[width=0.8\linewidth]{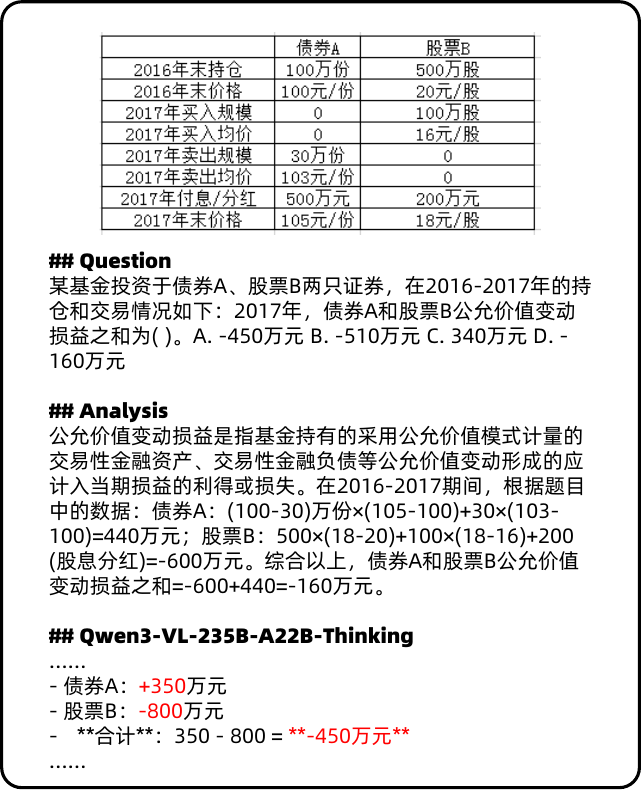}
    \end{minipage} 

\end{tcolorbox}
\caption{Visualization of error cases in LVLMs for our benchmarks. The red characters denote errors.}
\label{fig:error}
\end{figure}

\paragraph{Error Analysis.} Among the incorrect responses produced by the evaluated LVLMs, Figure \ref{fig:error} presents representative failure cases from the top model. As shown, the dominant error patterns include hallucinations, perceived barriers, insufficient understanding of finance-specific terminology, and irrational reasoning. These results demonstrate that even state-of-the-art models remain prone to systematic failures, and further confirm that our benchmark can comprehensively evaluate LVLMs’ capabilities in perception, understanding, reasoning, and cognition.

\paragraph{Cross-Modal Capability Analysis.} Joint understanding and reasoning over images and text constitutes core capabilities of LVLMs, yet they remain highly challenging, particularly for financial images with complex structures and high information density. We conduct experiments on table images from the question answering task in the application assessment, where these tables can be represented losslessly using HTML format. We compare the results obtained when using (i) a table image and (ii) the HTML of the table, together with the question as input. As shown in Table \ref{tab:cross}, we find that the latter yielded better performance. These results not only indicate that existing models remain inadequate at perceiving complex images accurately and comprehensively and aligning them with the linguistic modality, but also reflect the complexity and high information density of the images in our benchmark.

\paragraph{Multi-Orientation Analysis.} In real-world applications, images obtained by financial practitioners are often not captured in an upright orientation. Therefore, we evaluate the robustness of LVLMs in handling multi-orientated images, through rotating all images from the question answering task in the application assessment by 90, 180, and 270. As shown in Table \ref{tab:ori}, it can be observed that models perform much worse on multi-orientation images than on original images, showing a performance decrease of more than 10\%. These results emphasize the need for including more rotated samples in the training data to mitigate this challenge.

\begin{table}
    \centering
    \footnotesize
    \resizebox{\linewidth}{!}{
        \begin{tabular}{l|c|c}
        \toprule
        \textbf{Models} & \textbf{Table Image} & \textbf{Table HTML} \\
        \midrule
        Qwen3-VL-8B-Instruct & 59.09 & 66.67 \\
        Qwen3-VL-8B-Thinking & 65.90 & 76.51 \\
        Qwen3-VL-235B-A22B-Instruct & 61.36 & 75.76 \\
        Qwen3-VL-235B-A22B-Thinking & 68.18 & 78.03 \\
        \bottomrule
        \end{tabular}
    }
    \caption{Comparison of results obtained when using the table image and the HTML of the table together with the question as input.}
    \label{tab:cross}
\end{table}

\begin{table}
    \centering
    \footnotesize
    \resizebox{\linewidth}{!}{
        \begin{tabular}{l|c|c|c|c}
        \toprule
        \textbf{Models} & \textbf{$0^\circ$} & \textbf{$90^\circ$} & \textbf{$180^\circ$} & \textbf{$270^\circ$} \\
        \midrule
        Qwen3-VL-8B-Instruct & 43.80 & 31.01 & 32.53 & 31.77 \\
        Qwen3-VL-8B-Thinking & 44.22 & 31.77 & 31.39 & 30.89 \\
        Qwen3-VL-235B-A22B-Instruct & 52.57 & 42.53 & 41.90 & 42.15 \\
        Qwen3-VL-235B-A22B-Thinking & 54.94 &  41.27 & 43.29 & 43.04 \\
        \bottomrule
        \end{tabular}
    }
    \caption{Comparison of results on multi-oriented images.}
    \label{tab:ori}
\end{table}

\section{Conclusion}
This paper introduces CFMME, a novel Chinese multimodal evaluation benchmark dataset with 6,052 instances designed to evaluate the capabilities of LVLMs in the financial domain. We hope that CFMME will spur further progress in LVLMs, especially by improving their performance on Chinese financial multimodal tasks.

\section*{Acknowledgments}
We thank the anonymous reviewers and the area chair for their constructive comments. The authors of this paper were supported by Qwen DianJin Team\footnote{\url{https://github.com/aliyun/qwen-dianjin}} from Alibaba Cloud Computing.

\section*{Ethics}
CFMME is sourced from online textbooks, examinations, documents, websites, applications, and existing shared datasets, annotated by our professional annotators. We have secured permissions for the inclusion of each dataset in CFMME. It is important to note that all datasets within CFMME carries low ethical risks, with stringent measures in place to ensure the absence of any sensitive or personally identifiable information.

\section*{Limitations}
While CFMME has made a significant progress in evaluating LVLMs in the financial domain, it still has some limitations. Firstly, we use metrics such as accuracy, CDM, and TEDS to evaluate the performance, which may not provide a comprehensive assessment of LVLMs' outputs. Secondly, focusing on the zero-shot setting when preparing prompts may constrain LVLMs’ output quality, preventing them from performing as effectively as they do in a few-shot setting. Lastly, although CFMME covers various types of financial images, the types can be further expanded and refined in real-world.


\bibliography{custom}

\appendix

\section{Models}
\label{app:model}
Table~\ref{tab:vlms} provides more details of the LVLMs we evaluated, including the size of the model's parameters, mode, access method and creator.

\section{Prompts}
\label{app:prompt}
Figure~\ref{fig:prompt_1} to Figure~\ref{fig:prompt_7} illustrate the evaluation prompts used in the experiments.

\section{Parameters}
\label{app:para}
Table \ref{tab:para} provides more details of parameters used in the inference stage, including temperature, 
top\_p and presence\_penalty. All other parameters are set to default.

\section{Detailed Performance}
For the knowledge assessment, Table~\ref{tab:s_qa} shows the performance in accuracy and weighted F1 per subset on the question answering task. For the application assessment, Table~\ref{tab:ie} shows the performance in F1 score per source on the information extraction task, while Table~\ref{tab:det} shows the performance in mAP per category on the detection task.

\section{Examples}
\begin{figure*}[h]
\begin{tcolorbox}[title=Example, colback=gray!5, colframe=black, fonttitle=\bfseries]

\begin{minipage}{\linewidth}
\centering
\includegraphics[width=0.8\linewidth]{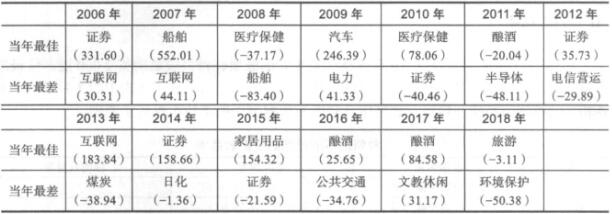}
\end{minipage}

\vspace{1em}

\begin{minipage}{\linewidth}
\begin{verbatim}
## Subject
投资学原理及应用
Principles and Applications of Investment
## Question
下表统计了中国股市2006～2018年涨跌幅度最大的行业，投资人据此主要可以体会四大投资准则中( )的重要性。
The table below summarizes the industries in China’s stock market with the 
largest gains and losses from 2006 to 2018. Based on this, investors can 
mainly appreciate the importance of which of the following four investment 
principles?
A. 尊重市场、适应市场
A. Respect the market and adapt to the market
B. 投资收益和投资风险形影相随
B. Investment returns and investment risks always go hand in hand
C. 牛熊市周而复始
C. Bull and bear markets occur in cycles
D. 分散投资降低风险
D. Diversification reduces risk
## Answer
D
## Analysis
无法从图中找到不同行业股票涨跌规律，只能分散投资。
It is impossible to identify consistent patterns of stock price rises and falls 
across different industries from the chart, so diversification is necessary.
\end{verbatim}
\end{minipage}

\end{tcolorbox}
\caption{An example in subject subset of the knowledge assessment.}
\label{fig:example_1}
\end{figure*}

\begin{figure*}[h]
\begin{tcolorbox}[title=Example, colback=gray!5, colframe=black, fonttitle=\bfseries]

\begin{minipage}{\linewidth}
\centering
\includegraphics[width=0.8\linewidth]{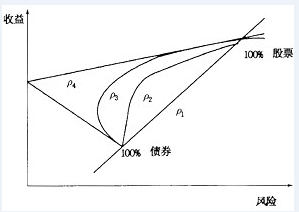}
\end{minipage}

\vspace{1em}

\begin{minipage}{\linewidth}
\begin{verbatim}
## Subject
证券专项考试
Specialized Securities Examination
## Question
资产组合的收益—风险特征如图所示，下列说法错误的是( )。
The return-risk characteristics of an asset portfolio are shown in the figure. 
Which of the following statements is incorrect?
A. p4对应的资产组合能最有效地降低风险
A. The portfolio corresponding to p4 can reduce risk most effectively.
B. 组合风险大小排序为：p4p3p2p1
B. The ranking of portfolio risk from highest to lowest is: p4, p3, p2, p1.
C. p1对应的资产组合不可能降低风险
C. The portfolio corresponding to p1 cannot reduce risk.
D. p1表示组成组合的两个资产不相关
D. p1 indicates that the two assets forming the portfolio are uncorrelated.
## Answer
D
## Analysis
p1表示组成组合的两个资产完全正相关，完全正相关的资产组成的组合无法分散风险。
p1 indicates that the two assets in the portfolio are perfectly positively 
correlated. A portfolio composed of perfectly positively correlated assets 
cannot diversify risk.
\end{verbatim}
\end{minipage}
\end{tcolorbox}
\caption{An example in the certification subset of the knowledge assessment.}
\label{fig:example_2}
\end{figure*}

\begin{figure*}[h]
\begin{tcolorbox}[title=Example, colback=gray!5, colframe=black, fonttitle=\bfseries]
\begin{minipage}[h]{0.48\textwidth}
\centering
\includegraphics[width=\linewidth]{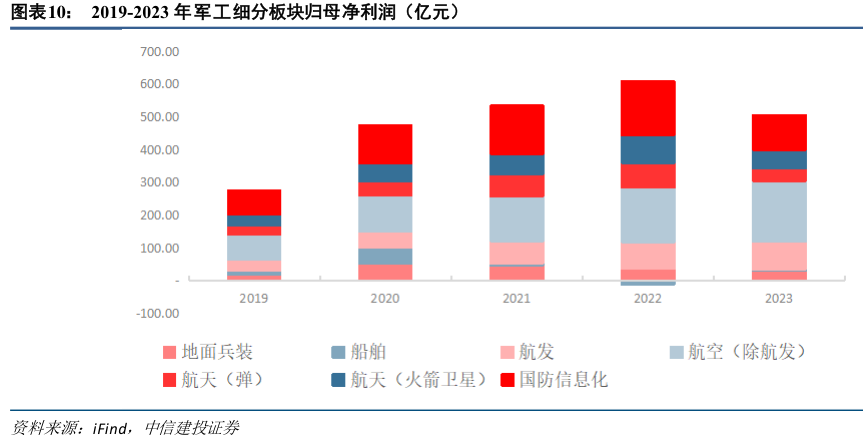}
\end{minipage}\hfill
\begin{minipage}[h]{0.48\textwidth}
\centering
\begin{verbatim}
## Question
2019-2023年里，军工细分板块里，哪一年某一细分板块亏损？（请输出年份；若无，则输出无）
During 2019–2023, in which year did any 
sub-sector within the defense industry 
incur a loss? (Please output the year; 
if none, output “None”.)
## Answer
2022
\end{verbatim}
\end{minipage}
\end{tcolorbox}
\caption{An example of the question answering task in the application assessment.}
\label{fig:example_3}
\end{figure*}

\begin{figure*}[h]
\begin{tcolorbox}[title=Example, colback=gray!5, colframe=black, fonttitle=\bfseries]
\begin{minipage}[h]{0.48\textwidth}
\centering
\includegraphics[width=0.5\linewidth]{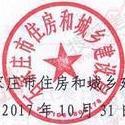}
\end{minipage}\hfill
\begin{minipage}[h]{0.48\textwidth}
\begin{verbatim}
## Answer
石家庄市住房和城乡建设局
\end{verbatim}
\end{minipage}
\end{tcolorbox}
\caption{An example of the seal recognition task in the application assessment.}
\label{fig:example_4}
\end{figure*}

\begin{figure*}[h]
\begin{tcolorbox}[title=Example, colback=gray!5, colframe=black, fonttitle=\bfseries]
\begin{minipage}{\linewidth}
\centering
\includegraphics[width=\linewidth]{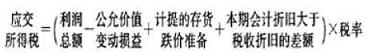}
\end{minipage}
\vspace{1em}
\begin{minipage}{\linewidth}
\begin{verbatim}
## Answer
应交所得税 = \left(利润总额 - 公允价值变动损益 + 计提的存货跌价准备 + 本期会计折旧大于税收折旧的差额 \right) \times 税率
\end{verbatim}
\end{minipage}
\end{tcolorbox}
\caption{An example of the formula recognition task in the application assessment.}
\label{fig:example_5}
\end{figure*}

\begin{figure*}[h]
\begin{tcolorbox}[title=Example, colback=gray!5, colframe=black, fonttitle=\bfseries]
\begin{minipage}{\linewidth}
\centering
\includegraphics[width=\linewidth]{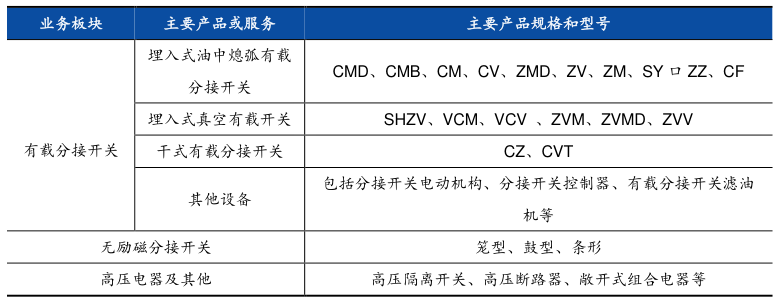}
\end{minipage}
\vspace{1em}
\begin{minipage}{\linewidth}
\begin{verbatim}
## Answer
<table>
    <tr>
        <td>业务板块</td>
        <td>主要产品或服务</td>
        <td>主要产品规格和型号</td>
    </tr>
    <tr>
        <td rowspan="4">有载分接开关</td>
        <td>埋入式油中熄弧有载分接开关</td>
        <td>CMD、CMB、CM、CV、ZMD、ZV、ZM、SY 口 ZZ、CF</td>
    </tr>
    <tr>
        <td>埋入式真空有载开关</td>
        <td>SHZV、VCM、VCV、ZVM、ZVMD、ZVV</td>
    </tr>
    <tr>
        <td>干式有载分接开关</td>
        <td>CZ、CVT</td>
    </tr>
    <tr>
        <td>其他设备</td>
        <td>包括分接开关电动机构、分接开关控制器、有载分接开关滤油机等</td>
    </tr>
    <tr>
        <td colspan="2">无励磁分接开关</td>
        <td>笼型、鼓型、条形</td>
    </tr>
    <tr>
        <td colspan="2">高压电器及其他</td>
        <td>高压隔离开关、高压断路器、敞开式组合电器等</td>
    </tr>
</table>
\end{verbatim}
\end{minipage}
\end{tcolorbox}
\caption{An example of the table recognition task in the application assessment.}
\label{fig:example_6}
\end{figure*}

\begin{figure*}[h]
\begin{tcolorbox}[title=Example, colback=gray!5, colframe=black, fonttitle=\bfseries]
\begin{minipage}[h]{0.48\textwidth}
\centering
\includegraphics[width=\linewidth]{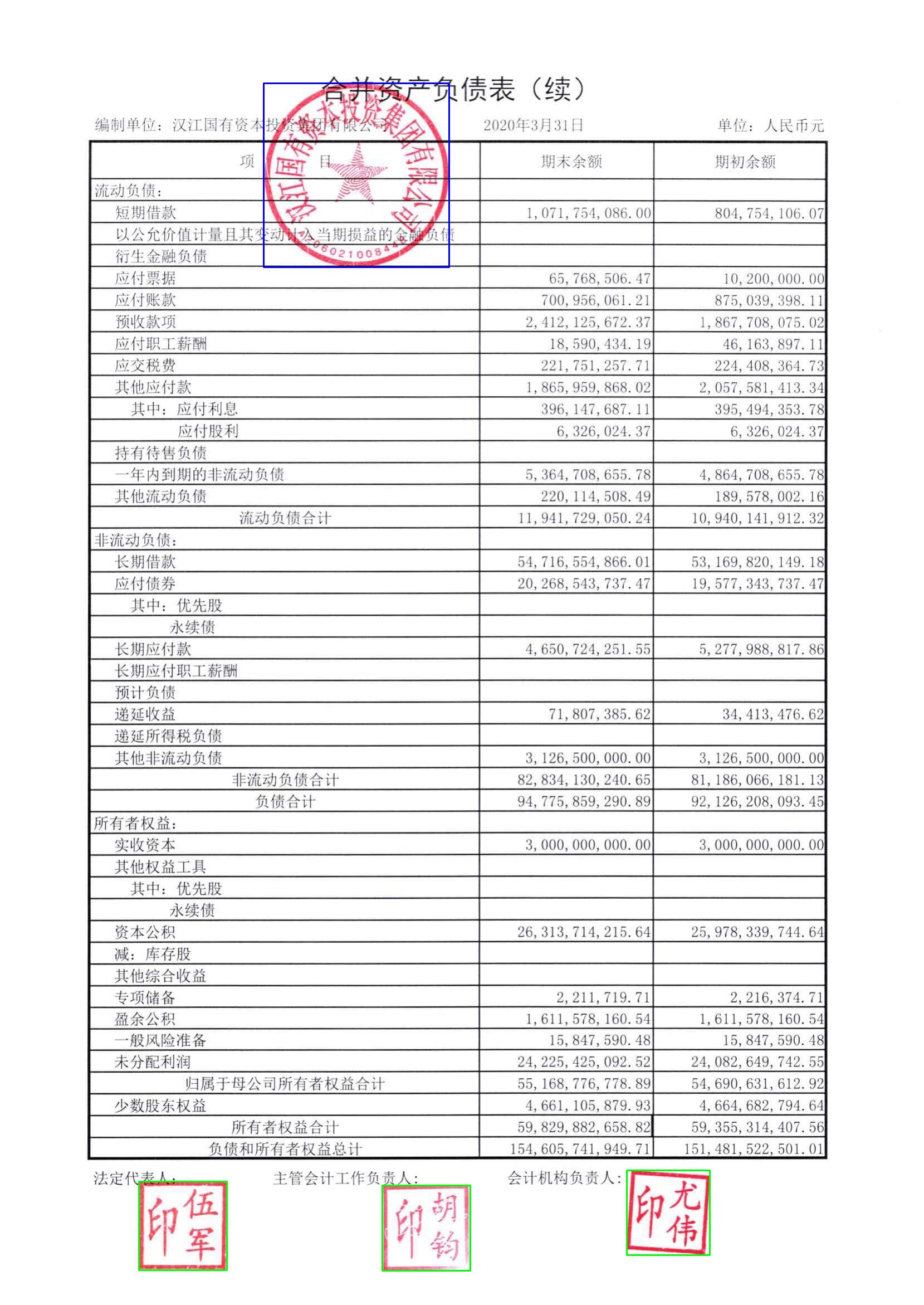}
\end{minipage}\hfill
\begin{minipage}[h]{0.48\textwidth}
\begin{verbatim}
## Answer
[{
    "bbox_2d": [473, 149, 806, 479],
    "label": "company"
},
{
    "bbox_2d": [249, 2120, 407, 2279],
    "label": "personal"
},
{
    "bbox_2d": [686, 2127, 844, 2280],
    "label": "personal"
},
{
    "bbox_2d": [1125, 2100, 1274, 2252],
    "label": "personal"
}]
\end{verbatim}
\end{minipage}

\end{tcolorbox}
\caption{An example of the detection task in the application assessment.}
\label{fig:example_7}
\end{figure*}

\begin{figure*}[h]
\begin{tcolorbox}[title=Example, colback=gray!5, colframe=black, fonttitle=\bfseries]
\begin{minipage}[h]{0.48\textwidth}
\centering
\includegraphics[width=\linewidth]{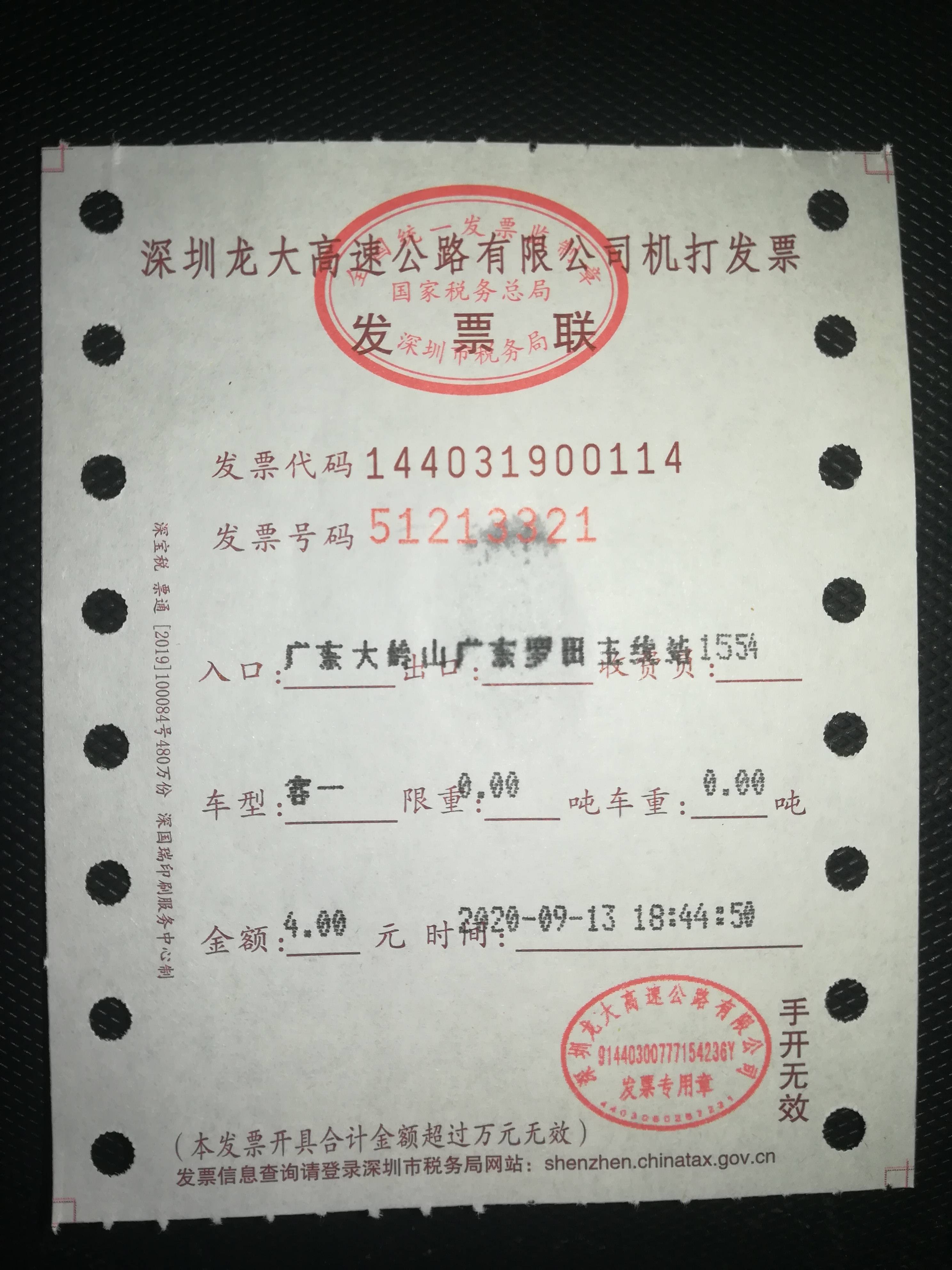}
\end{minipage}\hfill
\begin{minipage}[h]{0.48\textwidth}
\begin{verbatim}
## Answer
{
    "票据名称": "深圳龙大高速公路有限公司通行费机打发票",
    "发票代码": "144031900114",
    "发票号码": "51213321",
    "入口": "广东大岭山",
    "出口": "广东罗阳主线站",
    "收费员": "1554",
    "车型": "客一",
    "限重": "0.00吨",
    "吨车重": "0.00吨",
    "金额": "4.00元",
    "时间": "2020-09-13 18:44:50"
}
\end{verbatim}
\end{minipage}

\end{tcolorbox}
\caption{An example of the information extraction task in the application assessment.}
\label{fig:example_8}
\end{figure*}

Figure~\ref{fig:example_1} to Figure~\ref{fig:example_8} show some examples in our benchmark.

\begin{table*}
    \centering
    \footnotesize
    \begin{tabular}{l|c|c|c|c}
        \toprule
        \textbf{Models} & \textbf{\#Param.} & \textbf{Mode} & \textbf{Access} & \textbf{Creator} \\
        \midrule
        Kimi-VL-A3B-Instruct & 16B-A3B & Instruct & Weights & Moonshot AI \\
        Kimi-VL-A3B-Thinking & 16B-A3B & Thinking & Weights & Moonshot AI \\
        ERNIE-4.5-VL-28B-A3B & 28B-A3B & Instruct & Weights & Baidu \\
        ERNIE-4.5-VL-28B-A3B-Thinking & 28B-A3B & Thinking & Weights & Baidu \\
        GLM-4.1V-9B-Thinking & 9B & Thinking & Weights & Zhipu AI \\
        GLM-4.5V & 106B-A12B & Hybrid & Weights & Zhipu AI \\
        InternVL3.5-8B & 8B & Hybrid & Weights & Shanghai AI Lab \\
        InternVL3.5-241B-A28B & 241B-A28B & Hybrid & Weights & Shanghai AI Lab \\
        Qwen3-VL-8B-Instruct & 8B & Instruct & Weights & Alibaba \\
        Qwen3-VL-8B-Thinking & 8B & Thinking & Weights & Alibaba \\
        Qwen3-VL-235B-A22B-Instruct & 235B-A22B & Instruct & Weights & Alibaba \\
        Qwen3-VL-235B-A22B-Thinking & 235B-A22B & Thinking & Weights & Alibaba \\
        \midrule
        Llama-3.2-11B-Vision-Instruct & 11B & Instruct & Weights & Meta \\
        GPT-4o & UnKown & UnKown & API & OpenAI \\
        \bottomrule
    \end{tabular}
    \caption{Details of LVLMs benchmarked in this paper.}
    \label{tab:vlms}
\end{table*}

\begin{table*}[h]
    \centering
    \footnotesize
    \begin{tabular}{l|c|c|c}
        \toprule
        \textbf{Models} & \textbf{temperature} & \textbf{top\_p} & \textbf{presence\_penalty} \\
        \midrule
        Kimi-VL-A3B-Instruct & 0.2 & 1.0 & 0.0 \\
        Kimi-VL-A3B-Thinking & 0.8 & 1.0 & 0.0 \\
        ERNIE-4.5-VL-28B-A3B & 0.2 & 0.8 & 0.0 \\
        ERNIE-4.5-VL-28B-A3B-Thinking & 0.6 & 0.95 & 0.0 \\
        GLM-4.1V-9B-Thinking & 0.8 & 0.6 & 0.0 \\
        GLM-4.5V (non-thinking) & 1.0 & 0.0001 & 0.0 \\
        GLM-4.5V (thinking) & 1.0 & 0.0001 & 0.0 \\
        InternVL3.5-8B (non-thinking) & 0.8 & 0.8 & 0.0 \\
        InternVL3.5-8B (thinking) & 0.6 & 0.95 & 1.5 \\
        InternVL3.5-241B-A28B (non-thinking) & 0.8 & 0.8 & 0.0 \\
        InternVL3.5-241B-A28B (thinking) & 0.6 & 0.95 & 1.5 \\
        Qwen3-VL-8B-Instruct & 0.7 & 0.8 & 1.5 \\
        Qwen3-VL-8B-Thinking & 0.6 & 0.95 & 0.0 \\
        Qwen3-VL-235B-A22B-Instruct & 0.7 & 0.8 & 1.5  \\
        Qwen3-VL-235B-A22B-Thinking & 0.6 & 0.95 & 0.0 \\
        \midrule
        Llama-3.2-11B-Vision-Instruct & 0.6 & 0.9 & 0.0 \\
        GPT-4o & 1.0 & 1.0 & 0.0 \\
        \bottomrule
    \end{tabular}
    \caption{Parameters used in the inference stage.}
    \label{tab:para}
\end{table*}

\begin{CJK*}{UTF8}{gkai}

\begin{figure*}[h]
\begin{tcolorbox}[title=Prompt Template, colback=gray!5, colframe=black, fonttitle=\bfseries]
\begin{verbatim}
请回答以上的选择题，回答的最后一行必须遵循以下格式:
答案：\boxed{选项字母}（如\boxed{A}或\boxed{AB}等等）
Please answer the multiple-choice questions above. The last line of your response
must follow this format:
Answer: \boxed{option letter} (e.g., \boxed{A} or \boxed{AB}, etc.)
\end{verbatim}
\end{tcolorbox}
\caption{Prompt used in the knowledge assessment.}
\label{fig:prompt_1}
\end{figure*}

\begin{figure*}[h]
\begin{tcolorbox}[title=Prompt Template, colback=gray!5, colframe=black, fonttitle=\bfseries]
\begin{verbatim}
请回答上面的问题，最后将答案置于 \boxed{}.
Please answer the above question, and put the final answer in \boxed{}.
\end{verbatim}
\end{tcolorbox}
\caption{Prompt used for the question answering task in the application assessment.}
\label{fig:prompt_2}
\end{figure*}

\begin{figure*}[h]
\begin{tcolorbox}[title=Prompt Template, colback=gray!5, colframe=black, fonttitle=\bfseries]
\begin{verbatim}
请识别图中印章的文字内容。不需要其它解释，请严格按照如下格式直接输出：
Please recognize the text on the seal in the image. No other explanation is 
needed. Please output strictly in the following format:
\text{content here}
\end{verbatim}
\end{tcolorbox}
\caption{Prompt used for the seal recognition task in the application assessment.}
\label{fig:prompt_3}
\end{figure*}

\begin{figure*}[h]
\begin{tcolorbox}[title=Prompt Template, colback=gray!5, colframe=black, fonttitle=\bfseries]
\begin{verbatim}
请将图中的公式转成 LaTeX 格式。不需要其它解释，请严格按照如下格式直接输出：
Please convert the formula in the image into LaTeX format. No other explanation 
is needed. Please output strictly in the following format:
```latex
latex here
```
\end{verbatim}
\end{tcolorbox}
\caption{Prompt used for the formula recognition task in the application assessment.}
\label{fig:prompt_4}
\end{figure*}

\begin{figure*}[h]
\begin{tcolorbox}[title=Prompt Template, colback=gray!5, colframe=black, fonttitle=\bfseries]
\begin{verbatim}
请将图中的表格转成 HTML 格式。如果是跨页的表格，请先进行合并后输出。除表格以外的其它内容请不要输出。不需要其它解释，请严格按照如下格式直接输出：
Please convert the table in the image into HTML format. If the table spans multiple 
pages, please merge it first before outputting. Do not output anything other than 
the table. No other explanation is needed. Please output strictly in the following 
format:
```html
html here
```
\end{verbatim}
\end{tcolorbox}
\caption{Prompt used for the table recognition task in the application assessment.}
\label{fig:prompt_5}
\end{figure*}

\begin{figure*}[h]
\begin{tcolorbox}[title=Prompt Template, colback=gray!5, colframe=black, fonttitle=\bfseries]
\begin{verbatim}
请框选图中所有的公司印章和个人印章，严格按照以下的 JSON 格式输出：
Please locate bounding boxes of all company seals and personal seals in the 
image, and output strictly in the following JSON format:
```json
[
    {
        "bbox_2d": [x_min, y_min, x_max, y_max],
        "label": ""  // 公司印章或个人印章
    },
    ...
]
```
\end{verbatim}
\end{tcolorbox}
\caption{Prompt used for the detection task in the application assessment.}
\label{fig:prompt_6}
\end{figure*}

\begin{figure*}[h]
\begin{tcolorbox}[title=Prompt Template, colback=gray!5, colframe=black, fonttitle=\bfseries]
\begin{verbatim}
假设你是一位信息抽取专家，请抽取给定图像中的关键信息，并以 JSON 格式返回，不需要解释。如果某个字段不存在，请返回空字符串。请严格按照如下 JSON 格式输出：
You are an information extraction expert. Please extract the key information 
from the given image and return it in JSON format, with no explanation. If a 
field does not exist, return an empty string. Please output strictly in the 
following JSON format:
{schema}
\end{verbatim}
\end{tcolorbox}
\caption{Prompt used for the information extraction task in the application assessment.}
\label{fig:prompt_7}
\end{figure*}

\begin{table*}[h]
    \centering
    \footnotesize
    \resizebox{\textwidth}{!}{
        \begin{tabular}{l|cc|cc}
            \toprule
            \multirow{2}{*}{\textbf{Models}} & \multicolumn{2}{c|}{\textbf{Subject}} & \multicolumn{2}{c}{\textbf{Certification}} \\
            & Acc. & F1 & Acc. & F1 \\
            \midrule
            Kimi-VL-A3B-Instruct & 37.54±0.63 & 39.61±0.79 & 36.26±0.56 & 39.34±0.63 \\
            Kimi-VL-A3B-Thinking & 46.26±0.67 & 46.25±0.64 & 47.59±0.55 & 48.76±0.68 \\
            ERNIE-4.5-VL-28B-A3B & 48.16±0.51 & 48.00±0.50 & 49.56±0.38 & 50.07±0.36 \\
            ERNIE-4.5-VL-28B-A3B-Thinking & 64.35±1.22 & 66.23±1.31 & 58.17±0.18 & 62.65±0.16 \\
            GLM-4.1V-9B-Thinking & 56.88±1.05 & 58.37±0.80 & 62.43±0.40 & 63.38±0.46 \\
            GLM-4.5V (non-thinking) & 60.02±0.47 & 61.65±0.39 & 66.12±0.04 & 66.41±0.10 \\
            GLM-4.5V (thinking) & 68.51±1.19 & 70.94±1.02 & 70.69±0.40 & \underline{73.18±0.38} \\
            InternVL3.5-8B (non-thinking) & 53.68±1.30 & 52.81±1.27 & 53.27±0.26 & 52.78±0.26 \\
            InternVL3.5-8B (thinking) & 63.35±1.29 & 65.55±1.47 & 57.25±0.30 & 61.51±0.29 \\
            InternVL3.5-241B-A28B (non-thinking) & 63.70±0.67 & 63.00±0.75 & 64.14±0.41 & 62.96±0.45 \\
            InternVL3.5-241B-A28B (thinking) & 71.77±0.80 & 72.04±0.57 & 69.21±0.21 & 70.66±0.32 \\
            Qwen3-VL-8B-Instruct & 68.09±0.30 & 68.41±0.16 & 60.30±0.62 & 61.98±0.54 \\
            Qwen3-VL-8B-Thinking & 69.87±0.99 & 71.35±0.71 & 61.88±0.25 & 65.00±0.41 \\
            Qwen3-VL-235B-A22B-Instruct & \underline{76.69±0.63} & \underline{76.32±0.64} & \underline{73.39±0.62} & 73.13±0.56  \\
            Qwen3-VL-235B-A22B-Thinking & \textbf{80.66±0.59} & \textbf{81.05±0.86} & \textbf{76.42±0.76} & \textbf{77.06±0.35} \\
            \midrule
            Llama-3.2-11B-Vision-Instruct & 25.44±0.25 & 27.29±0.39 & 20.88±0.50 & 22.59±0.49 \\
            GPT-4o & 59.61±1.33 & 60.53±1.49 & 56.19±0.96 & 58.19±1.09 \\
            \midrule
        \end{tabular}
    }
    \caption{Performance of per subset in the knowledge assessment. Scores in \textbf{bold} and \underline{underline} denote the top and second-best performances.}
    \label{tab:s_qa}
\end{table*}

\begin{table*}[h]
  \centering
  \footnotesize
  \begin{tabular}{l|c|c|c}
  \toprule
  \textbf{Models} & \textbf{HUST-CELL} & \textbf{Baidu-FEST} & \textbf{CMB2017} \\
  \midrule
  Kimi-VL-A3B-Instruct & 75.66±1.44 & 79.82±0.24 & 49.33±0.38 \\
  ERNIE-4.5-VL-28B-A3B & 81.22±0.43 & 78.44±0.14 & 65.84±0.44 \\
  GLM-4.5V & 89.81±0.82 & 84.71±0.07 & \underline{68.07±0.09} \\
  InternVL3.5-8B & 81.36±0.69 & 72.51±0.25 & 59.73±0.81 \\
  InternVL3.5-241B-A28B & 83.68±0.21 & 80.14±0.09 & 67.07±0.84 \\
  Qwen3-VL-8B-Instruct & \underline{89.82±0.43} & \underline{84.89±0.05} & 58.80±0.49 \\
  Qwen3-VL-235B-A22B-Instruct & \textbf{94.01±0.39} & \textbf{90.88±0.09} & \textbf{75.20±0.59} \\
  \midrule
  Llama-3.2-11B-Vision-Instruct & 48.31±0.40 & 44.54±0.59 & 22.20±0.75 \\
  GPT-4o & 66.90±0.35 & 68.33±0.50 & 60.67±1.27 \\
  \bottomrule
  \end{tabular}
  \caption{Performance in F1 score per source on the information extraction task. Scores in \textbf{bold} and \underline{underline} denote the top and second-best performances.}
  \label{tab:ie}
\end{table*}

\begin{table*}[h]
    \centering
    \footnotesize
    \begin{tabular}{l|c|c}
        \toprule
        \textbf{Models} & \textbf{company} & \textbf{personal} \\
        \midrule
        GLM-4.5V & \textbf{54.86±0.27} & 19.33±0.07 \\
        Qwen3-VL-8B-Instruct & \underline{15.41±0.21} & \underline{40.78±0.24} \\
        Qwen3-VL-235B-A22B-Instruct & 10.58±0.21 & \textbf{49.28±0.11} \\
        \bottomrule
    \end{tabular}
    \caption{Performance in mAP per category on the detection task. Scores in \textbf{bold} and \underline{underline} denote the top and second-best performances.}
    \label{tab:det}
\end{table*}

\end{CJK*}
\end{document}